\documentclass[11pt]{article}
\pdfoutput=1

\usepackage{authblk}
\makeatletter
\renewcommand\maketitle{\AB@maketitle} 
\renewcommand\AB@affilsepx{\quad\protect\Affilfont} 
\makeatother

\usepackage{acl}

\usepackage{times}
\usepackage{latexsym}
\usepackage{adjustbox,lipsum}
\usepackage[T1]{fontenc}

\usepackage{arabtex}
\usepackage{utf8}
\setcode{utf8}
\usepackage{CJKutf8}

\usepackage{float}
\usepackage{booktabs}

\usepackage[utf8]{inputenc}

\usepackage{textcomp}
\usepackage{multirow}
\usepackage{graphicx}
\usepackage{xcolor}
\usepackage{array}
\usepackage{booktabs}
\definecolor{green}{RGB}{153,255,153}
\definecolor{hred}{RGB}{255,153,153}
\usepackage{soul}
\DeclareRobustCommand{\hlgreen}[1]{\sethlcolor{green}\hl{#1}}
\DeclareRobustCommand{\hlred}[1]{\sethlcolor{hred}\hl{#1}}
\newcommand{\PreserveBackslash}[1]{\let\temp=\\#1\let\\=\temp}
\newcolumntype{C}[1]{>{\PreserveBackslash\centering}p{#1}}

\usepackage{microtype}
\usepackage[suppress]{color-edits} 
\addauthor{ZT}{olive} 
\usepackage{multirow,tabularx,booktabs}
\everypar{\looseness=-1}

\title{\textsc{Multilingual HateCheck}: Functional Tests for\\ Multilingual Hate Speech Detection Models}

\author[1,2]{\textbf{Paul R\"ottger}}
\author[1]{\textbf{Haitham Seelawi}}
\author[1,3]{\textbf{Debora Nozza}}
\author[1,4]{\textbf{Zeerak Talat}}
\author[1]{\textbf{Bertie Vidgen}}
\affil[1]{Rewire}
\affil[2]{University of Oxford}
\affil[3]{Bocconi University}
\affil[4]{Simon Fraser University}

\begin{document}
\maketitle
\begin{abstracting}
Hate speech detection models are typically evaluated on held-out test sets.
However, this risks painting an incomplete and potentially misleading picture of model performance because of increasingly well-documented systematic gaps and biases in hate speech datasets.
To enable more targeted diagnostic insights, recent research has thus introduced functional tests for hate speech detection models.
However, these tests currently only exist for English-language content, which means that they cannot support the development of more effective models in other languages spoken by billions across the world.
To help address this issue, we introduce \textsc{Multilingual HateCheck} (MHC), a suite of functional tests for multilingual hate speech detection models.
MHC covers 34 functionalities across ten languages, which is more languages than any other hate speech dataset.
To illustrate MHC's utility, we train and test a high-performing multilingual hate speech detection model, and reveal critical model weaknesses for monolingual and cross-lingual applications.
\end{abstracting}
\section{Introduction} \label{sec:intro}
\vspace{-0.1cm}
Hate speech detection models play a key role in online content moderation and also enable scientific analysis and monitoring of online hate.
Traditionally, models have been evaluated by their performance on held-out test sets.
However, this practice risks painting an incomplete and misleading picture of model quality.
Hate speech datasets are prone to exhibit systematic gaps and biases due to how they are sampled \citep{wiegand2019detection,vidgen2020directions,poletto2021resources} and annotated \citep{waseem2016you,davidson2019racialbias,sap2021annotators}.
Therefore, models may perform deceptively well by learning overly simplistic decision rules rather than encoding a generalisable understanding of the task \citep[e.g.][]{niven2019probing,geva2019modeling,shah2020predictive}.
Further, aggregate and thus abstract performance metrics such as accuracy and F1 score may obscure more specific model weaknesses \citep{wu2019errudite}.

For these reasons, recent hate speech research has introduced novel test sets and methods that allow for a more targeted evaluation of model functionalities \citep{calabrese2021aaa,kirk2021hatemoji,mathew2021hatexplain,rottger2021hatecheck}.
However, these novel test sets, like most hate speech datasets so far, focus on English-language content.
A lack of effective evaluation hinders the development of higher-quality hate speech detection models for other languages.
As a consequence, billions of non-English speakers across the world are given less protection against online hate, and even the largest social media platforms have clear language gaps in their content moderation \citep{simonite2021facebook,marinescu2021facebook}.

As a step towards closing these language gaps, we introduce
\textsc{Multilingual HateCheck} (MHC), which extends the English \textsc{HateCheck} functional test suite for hate speech detection models \citep{rottger2021hatecheck} to ten more languages.
Functional testing evaluates models on sets of targeted test cases \citep{beizer1995black}.
\citet{ribeiro2020beyond} first applied this idea to structured model evaluation in NLP, and \citet{rottger2021hatecheck} used it to diagnose critical model weaknesses in English hate speech detection models.
We create novel functional test suites for Arabic, Dutch, French, German, Hindi, Italian, Mandarin, Polish, Portuguese and Spanish.\footnote{On dialects: we use Egyptian Arabic in Arabic script, European Dutch and French, High German, Standard Italian and Polish, Standard Hindi in Latin script, Standard Mandarin in Chinese script, Brazilian Portuguese and Argentinian Spanish.}
To our knowledge, MHC covers more languages than any other hate speech dataset.

The functional tests for each language in MHC broadly match those of the original \textsc{HateCheck}, which were selected based on interviews with civil society stakeholders as well as a review of hate speech research.
In each language, there are between 25 and 27 tests for different kinds of hate speech (e.g. dehumanisation and threatening language) as well as contrasting non-hate, which may lexically resemble hate speech but is clearly non-hateful (e.g. counter speech).
These contrasts make the test suites particularly challenging to models that rely on overly simplistic decision rules and thus enable more accurate evaluation of model functionalities \citep{gardner2020evaluating}.
For each functional test, native-speaking language experts hand-crafted targeted test cases with clear gold standard labels, using the English cases as a starting point but adapting them to retain realism and cultural compatibility in the target language.

We demonstrate MHC's utility as a diagnostic tool by evaluating a multilingual XLM-T model \citep{barbieri2021xlmt} fine-tuned on a combination of three widely-used hate speech datasets in Spanish, Italian and Portuguese.
This model achieves strong performance on the respective held-out test sets.
However, testing with MHC reveals that the model is 1) overly sensitive to key words and key phrases, 2) biased in its target coverage and 3) error-prone and inconsistent in cross-lingual transfer, in both zero- and many-shot settings.
If this model was used to moderate content, these critical weaknesses could cause serious harm, leaving some users unprotected from hate while others are restricted in their freedom of expression.
We hope that by revealing such weaknesses, MHC can play a key role in the development of better multilingual hate speech detection models.\footnote{We make MHC available \href{https://github.com/rewire-online/multilingual-hatecheck}{on GitHub}.}

\paragraph{Definition of Hate Speech}
\vspace{-0.1cm}
Following \citet{rottger2021hatecheck}, we define hate speech as \textit{abuse that is targeted at a protected group or at its members for being a part of that group}.
Protected groups are groups based on age, disability, gender identity, race, national or ethnic origins, religion, sex or sexual orientation, which broadly reflects Western legal consensus, particularly the US 1964 Civil Rights Act, the UK's 2010 Equality Act and the EU's Charter of Fundamental Rights.
Based on these definitions, we approach hate speech detection as the binary classification of content as either hateful or non-hateful.

\paragraph{Content Warning}
\vspace{-0.1cm}
This article contains examples of hateful and abusive language. All examples are taken from MHC in their original language, to illustrate MHC's composition.

\section{Constructing \textsc{MHC}}

\subsection{Defining Functional Tests}

Functional tests in MHC test the ability of hate speech detection models to correctly classify different types of hateful and non-hateful content.
The tests are fine-grained, distinct and contrastive.
For example, we have separate tests for hate expressed using profanity (e.g. DE:"F*ck alle Juden"\footnote{We use ISO 639-1 codes to denote the different languages when giving test case examples. See Table \ref{fig: f1_by_label} for a legend.}) and non-hateful uses of profanity (e.g. DE:"Ich hatte noch nie so einen guten F*ck") in every language.
Related functional tests are grouped into classes.

\subsection{Selecting Functional Tests} \label{subsec: func in test suite}

We selected functional tests for each language in MHC to broadly match those from the original \textsc{HateCheck}.
\newcite{rottger2021hatecheck}, in turn, motivated their selection of tests based on two factors:
1) a series of 21 interviews with NGO workers from the UK, US and Germany whose work directly relates to online hate, and
2) a review of previous hate speech research, particularly taxonomy work \citep[e.g.][]{zampieri2019predicting, banko2020unified, kurrek2020comprehensive}, error analyses \citep[e.g.][]{davidson2017automated, van2018challenges, vidgen2020detecting} and survey articles \citep[e.g.][]{schmidt2017survey, fortuna2018survey, vidgen2019challenges}.
All test cases are short text statements, and they are constructed to be clearly hateful or non-hateful according to our definition of hate speech.

Overall, there are 27 functional tests grouped into 11 classes for each of the ten languages in MHC, except for Mandarin, which has 25 functional tests.
Compared to the 29 functional tests in \textsc{HateCheck}, we 1) exclude slur homonyms and reclaimed slurs, because they have no direct equivalents in most MHC languages, and 2) adapt functional tests for spelling variations to non-Latin script in Arabic and Mandarin.
For Mandarin, there are two fewer tests for spelling variations and thus two fewer tests overall compared to the other nine languages.
As in \textsc{HateCheck}, the tests cover \textbf{distinct expressions of hate}, as well as \textbf{contrastive non-hate}, which shares lexical features with hate but is unambiguously non-hateful.
We provide example cases in different languages for each functional test in Appendix \ref{app: examples}.

\paragraph{Distinct Expressions of Hate}
MHC tests different types of derogatory hate speech (\textbf{F1-4}) and hate expressed through threatening language (\textbf{F5/6}).
It tests hate expressed using slurs (\textbf{F7}) and profanity (\textbf{F8}).
MHC also tests hate expressed through pronoun reference (\textbf{F10/11}), negation (\textbf{F12)} and phrasing variants, specifically questions and opinions (\textbf{F14/15}).
Lastly, MHC tests hate containing spelling variations such as missing characters or leet speak (\textbf{F23-34}), as well as spelling variations in non-Latin script for Arabic (\textbf{F28-31}) and Mandarin (\textbf{F32-34}).
For example, there is an Arabic-specific test for spellings in Arabizi, the Arabic chat alphabet (\textbf{F30}), and a Mandarin-specific test for spellings in Pinyin, Mandarin's romanised version (\textbf{F34}).

\paragraph{Contrastive Non-Hate}
MHC tests non-hateful contrasts which use profanity (\textbf{F9}) and negation (\textbf{F13}) as well as protected group identifiers (\textbf{F16/17}).
It also tests non-hateful contrasts in which hate speech is quoted or referenced, specifically \textit{counter speech}, i.e. direct responses to hate speech which seek to act against it (\textbf{F18/19}).
Lastly, MHC tests non-hateful contrasts which target out-of-scope entities such as objects (\textbf{F20-22}) rather than a protected group.

\subsection{Generating Test Cases} \label{subsec: generating test cases}
All test cases in MHC are hand-crafted by native-speaking language experts who have prior experience researching and/or annotating hate speech.\footnote{See Appendix \ref{app: data statement} for more detail on the language experts.}
Each test case is a short statement that corresponds to exactly one gold standard label.
\textsc{HateCheck}'s English test cases provide a starting point for MHC, but experts were encouraged to creatively adapt cases rather than providing literal translations, so as to retain relevance and realism.
Adapting language-specific idioms (e.g. "murder that beat"), slurs (e.g. "c*nt") and profanity (e.g. "f*ck") in particular required more creativity.

Test cases are generated at scale using templates \citep{dixon2018measuring,garg2019counterfactual,ribeiro2020beyond}, in which we replace tokens for protected group identifiers (e.g. NL:"Ik haat [IDENT].") and slurs (e.g. NL:"Voor mij ben je een [SLR].").
Compared to \textsc{HateCheck}, the templates for MHC required more granular placeholders for gender- and case-inflected languages.
German templates, for example, were instantiated based on gender and count of the identity group term, as well as its case: the male singular of Jew ("[male\_IDENT\_S]") in the German Akkusativ would be "den Juden" whereas its Nominativ would be "der Jude".
The benefits of the template approach are that
1) MHC has an equal number of cases targeted at different protected groups in each language, and
2) the templates can easily be used to generate more test cases targeted at other protected groups in the future.

MHC covers seven protected groups in each language.
The selection of protected groups varies by language, with the goal of maximising the relevance of MHC to the cultural context of each language.
Some groups, like women, are covered in all ten languages, whereas others, like lower caste people (HI:"shudra") are only covered in one language.
For cases targeted at or referencing a particular protected group, we provide a secondary label for the group that is targeted.
For an overview of protected groups covered by MHC in each language, see Appendix \ref{app: prot groups}.

In total, we generate 36,582 test cases across the ten languages in MHC, of which 25,511 ($69.7\%$) are hateful and 11,071 ($30.2\%$) are non-hateful.
There is some variation in the number of test cases between different languages due to different numbers of functional tests and slurs, with Polish having the most cases (3,815) and Mandarin the least (3,378).
88.6\% of all test cases are generated from templates.
The remaining test cases do not use template tokens (e.g. PT:"O clima hoje está f*da") and are crafted individually.
The average test case length is 8.5 words (std. dev. = 3.6) or 46.0 characters (std. dev. = 21.3).
This resembles the composition of the original English \textsc{HateCheck}.

\paragraph{Validation}
To validate gold standard labels in MHC, all 36,582 test cases were reviewed by three native-speaking annotators.
For each language we recruited at least 12 annotators, so that in total more than 120 annotators worked on MHC.\footnote{See Appendix \ref{app: data statement} for more detail on annotator recruitment.}
Annotation was prescriptive \citep{rottger2021two}, meaning that annotators were tasked with adhering to clear annotation guidelines reflecting our definition of hate speech, rather than sharing their personal view on what is and is not hateful.\footnote{Annotation guidelines are available \href{https://github.com/rewire-online/multilingual-hatecheck}{on GitHub}.}
Compared to the original \textsc{HateCheck}, where four out of five annotators confirmed the gold label for 99.4\% of all test cases, there was more disagreement on MHC (see Appendix \ref{app: disagreement}).
Upon review, language experts found that the vast majority of disagreement in MHC stemmed from annotators failing to apply the provided guidelines.
For example, annotators mislabelled several instances of clearly hateful slurs (e.g. DE:"Du bist eine F*tze, mehr nicht").
Therefore, we did not exclude any cases from MHC.
To enable further analysis and data filtering, we publish all annotator labels and mark up cases and templates where annotator majority labels differ from the gold labels.

\begin{table*}[]
\small
\centering
\renewcommand{\arraystretch}{1.33}
\resizebox{\textwidth}{!}{
    \begin{tabular}{p{0.08\textwidth}m{0.35\textwidth}C{0.1\textwidth}|C{0.03\textwidth}C{0.03\textwidth}C{0.03\textwidth}C{0.03\textwidth}C{0.03\textwidth}C{0.03\textwidth}C{0.03\textwidth}C{0.03\textwidth}C{0.03\textwidth}C{0.03\textwidth}}
    \toprule
     & \multirow{2}{*}{\textbf{Functionality}} & \multirow{2}{*}{\textbf{Gold Label}} & \multicolumn{10}{c}{\textbf{Accuracy (\%)}} \\
     & & & \textbf{AR} & \textbf{NL} & \textbf{FR} & \textbf{DE} & \textbf{HI} & \underline{\textbf{IT}} & \textbf{ZH} & \textbf{PL} & \underline{\textbf{PT}} & \underline{\textbf{ES}}  \\
    \midrule
    \multirow{1}{*}{\rotatebox[origin=c]{0}{\parbox[c]{1.3cm}{Derogation}}}
    & \textbf{F1}: Expression of strong negative emotions (explicit) & \hlred{ hateful } & 82.9 &   80.0 &    82.1 &    82.9 &   75.0 &     75.0 &      76.4 &    67.9 &        \textbf{83.6} &     80.7 \\
     & \textbf{F2}: Description using very negative attributes (explicit) & \hlred{ hateful } & 87.9 &   87.1 &    84.3 &    82.1 &   77.1 &     82.1 &      85.7 &    72.9 &        \textbf{91.4} &     78.6 \\
    & \textbf{F3}: Dehumanisation (explicit) & \hlred{ hateful } & 92.9 &   91.4 &    92.1 &    94.3 &   82.9 &     95.0 &      \textbf{97.9} &    87.9 &        91.4 &     84.3 \\
    & \textbf{F4}: Implicit derogation & \hlred{ hateful } & 75.2 &   72.1 &    74.3 &    63.4 &   \textbf{85.0} &     62.1 &      52.1 &    55.7 &        76.1 &     64.3 \\
    \midrule
    \multirow{2}{*}{\rotatebox[origin=c]{0}{\parbox[c]{1.2cm}{Threat. language}}}
    & \textbf{F5}: Direct threat & \hlred{ hateful } & 85.0 &   94.3 &    92.9 &    93.6 &   84.3 &     88.6 &      \textbf{97.9} &    76.4 &        90.0 &     89.3 \\
    & \textbf{F6}: Threat as normative statement & \hlred{ hateful } & 95.0 &   91.4 &    92.1 &    93.6 &   \textbf{96.4} &     91.4 &      92.1 &    84.3 &        90.7 &     91.0 \\
    \midrule
    Slurs
    & \textbf{F7}: Hate expressed using slur & \hlred{ hateful } & 76.9 &   55.3 &    73.8 &    67.5 &   64.4 &     52.1 &      70.7 &    51.1 &        \textbf{77.1} &    \textit{\textcolor{red}{43.3}} \\
    \midrule
    \multirow{2}{10pt}{\rotatebox[origin=c]{0}{\parbox[c]{1.2cm}{Profanity usage}}} 
    & \textbf{F8}: Hate expressed using profanity & \hlred{ hateful } & 94.3 &   83.6 &    91.0 &    90.0 &   80.0 &     80.0 &      \textbf{97.1} &    79.3 &        94.3 &     75.7 \\
    & \textbf{F9}: Non-hateful use of profanity & \hlgreen{ non-hate } & 61.0 &   91.0 &    77.0 &    91.0 &   57.0 &     79.0 &      74.0 &    92.0 &        79.0 &     \textbf{99.0} \\
    \midrule
    \multirow{1}{*}{\rotatebox[origin=c]{0}{\parbox[c]{1.3cm}{Pronoun reference}}}
    & \textbf{F10}: Hate expressed through reference in subsequent clauses & \hlred{ hateful } & 84.3 &   81.4 &    \textbf{94.0} &    89.3 &   90.7 &     91.4 &      83.6 &    65.0 &        86.4 &     84.1 \\
     & \textbf{F11}: Hate expressed through reference in subsequent sentences & \hlred{ hateful } & 88.6 &   90.0 &    \textbf{91.4} &    \textbf{91.4} &   89.3 &     87.9 &      89.3 &    69.3 &        85.0 &     79.3 \\
    \midrule
    \multirow{1}{*}{\rotatebox[origin=c]{0}{\parbox[c]{1.3cm}{Negation}}}
    & \textbf{F12}: Hate expressed using negated positive statement & \hlred{ hateful } & \textbf{89.3} &   67.9 &    72.1 &    70.0 &   87.9 &     72.9 &      72.1 &    65.0 &        82.1 &     67.6 \\
    & \textbf{F13}: Non-hate expressed using negated hateful statement & \hlgreen{ non-hate } & \textit{\textcolor{red}{17.9}} &   \textit{\textcolor{red}{33.6}} &    \textit{\textcolor{red}{27.9}} &    \textit{\textcolor{red}{28.6}} &   \textit{\textcolor{red}{10.0}} &     \textit{\textcolor{red}{33.6}} &      \textit{\textcolor{red}{43.6}} &    \textit{\textcolor{red}{35.0}} &        \textit{\textcolor{red}{19.3}} &     \textit{\textcolor{red}{35.7}} \\
    \midrule
    \multirow{2}{*}{\rotatebox[origin=c]{0}{\parbox[c]{1.3cm}{Phrasing}}}
    & \textbf{F14}: Hate phrased as a question & \hlred{ hateful } & 88.6 &   73.6 &    84.3 &    91.4 &   77.9 &     87.9 &      74.3 &    75.0 &        \textbf{93.6} &     72.9 \\
    & \textbf{F15}: Hate phrased as an opinion & \hlred{ hateful } & 90.7 &   77.9 &    \textbf{92.9} &    87.9 &   78.6 &     89.3 &      90.0 &    75.0 &        92.1 &     78.6 \\
    \midrule
    \multirow{2}{*}{\rotatebox[origin=c]{0}{\parbox[c]{1.3cm}{Non-hateful group identifier}}} 
    & \textbf{F16}: Neutral statements using protected group identifiers & \hlgreen{ non-hate } & \textit{\textcolor{red}{44.3}} &   67.1 &    67.1 &    67.9 &   \textit{\textcolor{red}{49.3}} &     83.6 &      80.0 &    \textbf{88.6} &        56.6 &     68.6 \\
    & \textbf{F17}: Positive statements using protected group identifiers & \hlgreen{ non-hate } & \textit{\textcolor{red}{47.6}} &   51.0 &    56.0 &    59.0 &   \textit{\textcolor{red}{40.0}} &     \textbf{81.4} &      75.7 &    73.8 &        67.1 &     71.4 \\
    \midrule
    \multirow{2}{*}{\rotatebox[origin=c]{0}{\parbox[c]{1.3cm}{Counter speech}}}
    & \textbf{F18}: Denouncements of hate that quote it & \hlgreen{ non-hate } & \textit{\textcolor{red}{10.3}} & \textit{\textcolor{red}{47.6}} &    \textit{\textcolor{red}{22.8}} &    \textit{\textcolor{red}{31.6}} &   \textit{\textcolor{red}{17.1}} &     \textit{\textcolor{red}{37.9}} &      \textit{\textcolor{red}{43.9}} &    \textbf{61.4} &        \textit{\textcolor{red}{22.4}} &     \textit{\textcolor{red}{40.9}} \\
    & \textbf{F19}: Denouncements of hate that make direct reference to it & \hlgreen{ non-hate } & \textit{\textcolor{red}{9.6}} &   \textit{\textcolor{red}{32.9}} &    \textit{\textcolor{red}{18.6}} &    \textit{\textcolor{red}{34.8}} &   \textit{\textcolor{red}{13.7}} &     \textit{\textcolor{red}{31.7}} &      \textit{\textcolor{red}{45.1}} &    \textit{\textcolor{red}{44.9}} &        \textit{\textcolor{red}{24.8}} &     \textbf{64.6} \\
    \midrule
    \multirow{4}{*}{\rotatebox[origin=c]{0}{\parbox[c]{1.3cm}{Abuse against non-protected targets}}}
    & \textbf{F20}: Abuse targeted at objects & \hlgreen{ non-hate } & 53.8 &   73.8 &    72.3 &    66.2 &   \textit{\textcolor{red}{49.2}} &     73.8 &      70.8 &    87.7 &        70.8 &     \textbf{86.2} \\
    & \textbf{F21}: Abuse targeted at individuals (not as member of a protected group) & \hlgreen{ non-hate } & 56.2 &   66.2 &    72.3 &    67.7 &   \textit{\textcolor{red}{41.5}} &     60.0 &      52.3 &    90.8 &        55.4 &     \textbf{92.3} \\
    & \textbf{F22}: Abuse targeted at non-protected groups (e.g. professions) & \hlgreen{ non-hate } & \textit{\textcolor{red}{27.7}} &   \textit{\textcolor{red}{47.7}} &    55.4 &    \textit{\textcolor{red}{33.8}} &   \textit{\textcolor{red}{26.2}} &     56.9 &      \textit{\textcolor{red}{38.5}} &    61.5 &        \textit{\textcolor{red}{44.6}} &     \textbf{70.8} \\
    \midrule
    \multirow{2}{*}{\rotatebox[origin=c]{0}{\parbox[c]{1.3cm}{Spelling variations}}}
    & \textbf{F23}: Swaps of adjacent characters & \hlred{ hateful } & - &   82.1 &    89.3 &    89.3 &   87.9 &     75.7 &       - &    69.3 &        \textbf{94.3} &     84.3\\
    & \textbf{F24}: Missing characters & \hlred{ hateful } & - &   85.0 &    75.0 &    82.9 &   74.3 &     69.3 &       - &    72.9 &        \textbf{85.7} &     68.6 \\
    & \textbf{F25}: Missing word boundaries & \hlred{ hateful } & - &   81.2 &    91.0 &    80.6 &   87.7 &     90.7 &       - &    71.6 &        \textbf{95.0} &     76.8 \\
    & \textbf{F26}: Added spaces between chars & \hlred{ hateful } & 65.8 &   61.2 &    86.8 &    85.8 &   \textbf{89.7}  &     77.0 &       - &    58.0 &        83.2 &     71.3 \\
    & \textbf{F27}: Leet speak spellings & \hlred{ hateful } & - & 94.7 &    \textbf{95.2} &    93.5 &   87.7 &     86.3 &       - &    71.6 &        95.0 &     81.1 \\
    & \textbf{F28}: AR: Latin char. replacement & \hlred{ hateful } & \textbf{83.0} &    - &     - &     - &    - &      - &       - &     - &         - &      - \\
    & \textbf{F29}: AR: Repeated characters & \hlred{ hateful } & \textbf{82.9} &    - &     - &     - &    - &      - &       - &     - &         - &      - \\
    & \textbf{F30}: AR: Arabizi (Arabic chat alphabet) & \hlred{ hateful } & \textbf{60.9} &    - &     - &     - &    - &      - &       - &     - &         - &      - \\
    & \textbf{F31}: AR: Accepted alt. spellings & \hlred{ hateful } & \textbf{85.6} &    - &     - &     - &    - &      - &       - &     - &         - &      - \\
    & \textbf{F32}: ZH: Homophone char. replacement & \hlred{ hateful } & - &    - &     - &     - &    - &      - &      \textbf{89.3} &     - &         - &      - \\
    & \textbf{F33}: ZH: Character decomposition & \hlred{ hateful } & - &    - &     - &     - &    - &      - &      \textbf{87.7} &     - &         - &      - \\
    & \textbf{F34}: ZH: Pinyin spelling & \hlred{ hateful } & - &    - &     - &     - &    - &      - &      \textbf{76.5} &     - &         - &      - \\
    \bottomrule
    \end{tabular}
}
\caption{\label{fig: functionalities}
\textsc{MHC} covers 34 functionalities in 11 classes with a total of n = 36,582 test cases.
69.74\% of cases (25,511 in 25 functional tests) are labelled \hlred{hateful}, 30.26\% (11,071 in 9 functional tests) are labelled \hlgreen{non-hateful}.
The right-most columns report accuracy (\%) of the the XTC model (\S\ref{subsec: model setup}) across functional tests for each language.
Languages which XTC was directly trained on are \underline{underlined}, to highlight many-shot vs. zero-shot settings.
Best performance on each functional test is \textbf{bolded}.
Below random choice performance ($<$50\%) is in \textit{\textcolor{red}{cursive red}}.
Examples of test cases for each functional test are listed in Appendix \ref{app: examples}.
}
\end{table*}
\section{Testing Models with \textsc{MHC}}

\subsection{Model Setup} \label{subsec: model setup}
As a suite of functional tests, \textsc{MHC} is broadly applicable across hate speech detection models for the ten languages that it covers.
Users can test multilingual models across all ten languages or use a language-specific test suite to test monolingual models.
MHC is model agnostic, and can be used to compare different architectures or different datasets in zero-, few- or many-shot settings, and even commercial models for which public information on architecture and training data is limited.

\paragraph{Multilingual Transformer Models}
We test XLM-T \citep{barbieri2021xlmt}, an XLM-R model \citep{conneau2020xlmr} pre-trained on an additional 198 million Twitter posts in over 30 languages.\footnote{We use the XLM-T implementation hosted on HuggingFace: \href{https://huggingface.co/cardiffnlp/twitter-xlm-roberta-base}{huggingface.co/cardiffnlp/twitter-xlm-roberta-base}.}
XLM-R is a widely-used architecture for multilingual language modelling, which has been shown to achieve near state-of-the-art performance on multilingual hate speech detection \citep{banerjee2021exploring,mandl2021hasoc}.
We chose XLM-T over XLM-R after initial experiments showed the former to outperform the latter on several hate speech detection datasets as well as MHC.

We fine-tune XLM-T on three widely-used hate speech datasets -- one Spanish \citep{basile2019semeval}, one Italian \citep{sanguinetti2020haspeede} and one Portuguese \citep{fortuna2019hierarchically}.
Accordingly, model performance is many-shot for Spanish, Italian and Portuguese, and zero-shot for all other languages.

All three datasets have an explicit label for hate speech that matches our definition of hate (\S\ref{sec:intro}), so that we can collapse all other labels into a single non-hateful label, to match MHC's binary format.
The Spanish \citet{basile2019semeval} dataset contains 4,950 tweets, of which 41.5\% are labelled as hateful.
The Italian \citet{sanguinetti2020haspeede} dataset contains 8,100 tweets, of which 41.8\% are labelled as hateful.
The Portuguese \citet{fortuna2019hierarchically} dataset contains 5,670 tweets, of which 31.5\% are labelled as hateful.

We focus our discussion on XTC, an XLM-T model fine-tuned on a combination of these three datasets, which outperforms XLM-T models fine-tuned on the three datasets individually (see Appendix \ref{app: model comparison}).
For the Spanish and Portuguese data, we use stratified 80/10/10 train/dev/test splits.
For the Italian data, we use the original 91.6/8.4 train/test split, and then split the original training set into 90/10 train/dev portions.
On the held-out test sets, XTC achieves 84.7 macro F1 for Spanish, 76.3 for Italian, and 73.3 for Portuguese, which is better than results reported in the original papers.\footnote{See Appendix \ref{app: training data} for details on each dataset and pre-processing, and Appendix \ref{app: hyperparameters} for details on model training.}

\paragraph{Testing Commercial Models}
Few commercial models for hate speech detection are available for research use, and only a small subset of them can handle non-English language content.
The best candidate for testing is Perspective, a free API built by Google's Jigsaw team.\footnote{\href{https://www.perspectiveapi.com/}{www.perspectiveapi.com/}}
Given an input text, Perspective provides percentage scores for attributes such as ``toxicity'' and ``identity attack''.
The "toxicity" attribute covers a wide range of languages, including the ten in MHC.
However, compared to hate speech, "toxicity" is a much broader concept, which includes other forms of abuse and profanity -- some of which would be considered contrastive non-hate in the context of MHC.
On the other hand, Perspective's ``identity attack'' aims to identify ``negative or hateful comments targeting someone because of their identity'' and thus aligns with our definition of hate speech (\S\ref{sec:intro}), but it is only available for three languages in MHC -- German, Italian and Portuguese.
For these three languages, XTC consistently outperforms Perspective (see Appendix \ref{app: perspective results}).

\subsection{Results}

\paragraph{Performance Across Labels}
MHC reveals clear gaps in XTC's performance across all ten languages (Table \ref{fig: f1_by_label}).
Overall performance in terms of macro F1 is best on Mandarin (71.5), Italian (69.6) and Spanish (69.5), and worst on Hindi (58.1), Arabic (59.4) and Polish (66.2).
F1 scores are higher for hateful cases than for non-hateful cases across all languages, with Hindi and Arabic exhibiting the biggest differences between hate and non-hate ($\sim$40pp).
For hateful cases, XTC performs best in terms of F1 score on Portuguese (83.5) and worst on Polish (76.1), but performance differences are relatively small across languages (< 8pp).
For non-hateful cases, on the other hand, performance varies considerably across languages (< 24pp), with XTC performing best on Mandarin (61.1) and worst on Hindi (39.8).

\begin{table}[H]
\centering
\begin{tabular}{l|cc|c}
\toprule
  \textbf{Language} & \textbf{F1-h} & \textbf{F1-nh} & \textbf{Mac. F1} \\
\midrule
    \textbf{Arabic / AR} &   79.1 &       39.8 &    59.4 \\
    \textbf{Dutch / NL} &   80.1 &       53.3 &    66.7 \\
    \textbf{French / FR} &   82.6 &       52.6 &    67.6 \\
    \textbf{German / DE} &   82.6 &       55.2 &    68.9 \\
    \textbf{Hindi / HI} &   78.5 &       37.7 &    58.1 \\
    \underline{\textbf{Italian / IT}} &   81.5 &       57.8 &    69.6 \\
    \textbf{Mandarin / ZH} &   81.8 &       61.1 &    71.5 \\
    \textbf{Polish / PL} &   76.1 &       56.4 &    66.2 \\
    \underline{\textbf{Portuguese / PT}} &   83.5 &       53.4 &    68.5 \\
    \underline{\textbf{Spanish / ES}} &   79.9 &       59.1 &    69.5 \\
\bottomrule
\end{tabular}
\caption{\label{fig: f1_by_label}
Performance of XTC across the ten languages in MHC. Many-shot settings are \underline{underlined}. All other languages are zero-shot.  F1 score for \textbf{h}ateful and \textbf{n}on-\textbf{h}ateful cases, and overall macro F1 score.}
\end{table}

\paragraph{Performance Across Functional Tests}
Evaluating XTC on each functional test across languages (Table \ref{fig: functionalities}) reveals specific model weaknesses.

XTC performs better than a random binary-choice baseline (50\% accuracy) on all functional tests for hate, with the exception of Spanish statements with hateful slurs (\textbf{F7}, 43.3\% accuracy).
Explicit dehumanisation (\textbf{F3}), threatening language (\textbf{F5/6}) and hate expressed using profanity (\textbf{F8}) appear to be the least challenging for the model, with relatively high and consistent accuracy across languages.
In comparison, XTC generally performs worse on implicit hate (\textbf{F4}) and spelling variations (\textbf{F23+}).
For other hateful functional tests, performance differs noticeably between languages.
For example, XTC is very accurate on \textbf{F10}: hate expressed through reference in subsequent clauses in Spanish (94.0\%), but much less so on Polish hate of the same kind (65.0\%).
Performance is worst on hate expressed using slurs (\textbf{F7}), with XTC misclassifying Spanish (43.3\%), Polish (51.1\%) and Italian statements (52.1\%) at particularly high rates.

The only non-hateful functional test where performance is better than a random choice baseline across all ten languages is for non-hateful uses of profanity (\textbf{F9}).
In contrast, negated hate (\textbf{F13}) and counter speech (\textbf{F18/19}), are misclassified at very high rates in every language.
For example, XTC is only 9.6\% accurate on Arabic denouncements of hate that make direct reference to it (\textbf{F22}).
Performance differs more noticeably across languages for non-hateful uses of protected group identity terms (\textbf{F16/17}) and abuse against non-protected targets (\textbf{F20-22}).
For example, XTC is 81.4\% accurate on positive statements using group identifiers in Italian, but only 40.0\% accurate on the same type of statement in Hindi.

\paragraph{Performance Across Target Groups}
MHC can show whether models exhibit \textit{unintended biases} in target coverage \citep{dixon2018measuring} by testing their performance on cases generated from [IDENT] templates.
These cases are exactly the same, apart from the group that is targeted (e.g. IT: "Odio i gay" and IT:"Odio gli immigrati").
To illustrate, Table \ref{fig: target coverage} shows model accuracy on these cases for French, Italian and Mandarin, which all cover the same seven protected groups.


\begin{table}[h]
    \centering
    
    \begin{tabular}{l|ccc}
\toprule
   \textbf{Target Group} &  \textbf{FR} &  \textbf{IT} & \textbf{ZH}  \\
\midrule
          Women & 70.9 & 70.1 &  74.5 \\
   Trans people &      65.4 & 67.6 & 71.0  \\
     Gay people &      56.4 & 68.2 & 70.3  \\
   Black people &      61.5 & 67.7 & 66.7  \\
Disabled people &      61.2 & 64.4 & 72.6  \\
        Muslims &    66.2 & 69.0 & 68.3  \\
     Immigrants &    66.8 & 72.4 & 74.0  \\

\bottomrule
\end{tabular}
\caption{Macro F1 of XTC on French, Italian and Mandarin test cases in MHC generated from [IDENT] templates, split by targeted protected group.}
\label{fig: target coverage}
\end{table}

In French, XTC performs substantially worse on cases targeting gay people (56.4 macro F1) than on equivalent cases targeting women (70.9) or immigrants (66.8).
In Italian, XTC performs particularly poorly on cases targeting disabled people (64.4), and in Mandarin on cases targeting black people (66.7), compared to other target groups.

\subsection{Discussion}
MHC reveals clear functional weaknesses in the XTC model across all ten languages.

First, XTC is overly sensitive to key words and phrases.
XTC does have some ability to distinguish between different uses of potentially hateful key words, as demonstrated by its relatively good performance on both hateful and non-hateful uses of profanity across languages (\textbf{F8/9}).
However, in languages like Hindi and Arabic, even entries which refer positively to protected groups (\textbf{F17}) are misclassified as hateful at high rates.
Accuracy on non-hateful contrasts to hateful phrases, like negated hate (\textbf{F13}) and counter speech (\textbf{F18/19}), is low across all languages.
Thus, XTC appears to rely at least in part on overly simplistic decision rules (e.g. that the use of group identifiers is a signal for hate speech).
It does not sufficiently register linguistic signals like negation, that reframe hateful phrases into clearly non-hateful statements (e.g. FR:"Je ne déteste pas les blacks.").

Second, XTC is biased in its target coverage across languages, classifying hate directed against some protected groups less accurately than equivalent cases directed at others (Table~\ref{fig: target coverage}).

These are clear weaknesses that were not apparent from XTC's strong performance on held-out test sets.
They match findings from \textsc{HateCheck} for monolingual English models \citep{rottger2021hatecheck}.
If XTC was used to moderate content, these weaknesses could cause serious harm.
In particular, misclassifying counter speech risks undermining positive efforts to fight hate speech, and 
biased target coverage may create and entrench biases in the protections afforded to different groups.
However, the multilingual nature of MHC also allows for additional, novel insights.

First, we can evaluate cross-lingual performance in both zero- and many-shot settings (Table \ref{fig: f1_by_label}).
XTC performs particularly well on Italian, Spanish and Portuguese -- the languages it was fine-tuned on -- but also on French, which is another Romance language.
Performance on other European languages is also relatively high.
By contrast, Hindi and Arabic clearly stand out as particularly challenging, with substantially lower performance. 
This suggests that cross-lingual transfer works better across more closely related languages and poses a challenge for more dissimilar languages.\footnote{The surprisingly good performance of XTC on Mandarin is a caveat, which may in part be explained by Mandarin being more prevalent than Arabic or Hindi in XLM-R's pre-training corpus \citep{conneau2020xlmr}.}
Cultural differences across language settings may also affect transferability.
We may for example expect hate in Italian and French to be more similar to each other than to hate in Hindi, along such dimensions as who the targets of hate are, which would likely affect the cross-lingual performance of hate speech detection models.
Both hypotheses could be explored in future research.

Second, we can evaluate differences in language-specific model behaviour, again in zero- as well as many-shot settings.
For example, XTC tends to overpredict hate in Hindi and Arabic, both zero-shot, whereas it tends to underpredict hate in many-shot Spanish and zero-shot Polish (Table~\ref{fig: functionalities}).
XTC also exhibits different target biases across languages, for zero-shot settings like in French and Mandarin as well as many-shot Italian (Table~\ref{fig: target coverage}).
This suggests that, in addition to accounting for differences in high-level performance, multilingual models may require very different calibration and adaptation across languages, even for languages they were not directly fine-tuned on.

Overall, the insights generated by MHC suggest two potential steps towards the development of more effective multilingual hate speech detection models:
1) creating training data in diverse languages to reduce language gaps, even for models with significant cross-lingual transfer abilities, and 
2) evaluating and addressing language-specific model biases as well as differences in performance across languages.
\section{Limitations} \label{sec:limitations}
\vspace{-0.1cm}
The limitations of the original \textsc{HateCheck} also apply to MHC.
First, MHC diagnoses specific model weaknesses rather than generalisable model strengths, and should be used to complement rather than substitute evaluation on held-out test sets of real-world hate speech.
Second, MHC does not test functionalities related to context outside of individual documents or modalities other than text.
Third, MHC only covers a limited set of protected groups and slurs across languages, but can easily be expanded using the provided case templates.

The multilingual nature of MHC creates additional considerations.
First, comparisons of performance between languages are not strictly like-for-like, because cases in different languages are not literal translations of each other.
This limitation is compounded for Arabic and Mandarin, which have unique functional tests for spelling variations.
Second, even though MHC includes a diverse set of ten languages, these languages still only make up a fraction of languages spoken across the world.
To our knowledge, MHC covers more languages than any other hate speech dataset, but hundreds of other languages remain neglected and should be considered for future expansions of MHC.
Third, the selection of functional tests in MHC is based on \textsc{HateCheck}, which was informed in part by interviews in an anglo-centric setting.
We worked with native-speaking language experts and created additional tests to account for non-Latin scripts in Arabic and Mandarin, but future research may consider additional interviews or other language-specific steps to inform expansions of MHC.
Lastly, individual languages, like the ten included in MHC, are not monolithic but vary between speakers, especially across geographic regions and sociodemographic groups.
We use widely-spoken dialects for the ten languages in MHC (see \S\ref{sec:intro}), but cannot cover all variations.
\section{Related Work} \label{sec:relatedwork}
\vspace{-0.1cm}
\paragraph{Diagnostic Hate Speech Datasets}
The concept of functional testing from software engineering \citep{beizer1995black} was first applied to NLP model evaluation by \citet{ribeiro2020beyond}.
The original \textsc{HateCheck} \citep{rottger2021hatecheck} then introduced functional tests for hate speech detection models, using hand-crafted test cases to diagnose model weaknesses on different kinds of hate and non-hate.
\citet{kirk2021hatemoji} applied the same framework to emoji-based hate.
\citet{manerba2021abusechecklist} provide smaller-scale functional test for abuse detection systems.
Other research has instead collected real-world examples of hate and annotated them for more fine-grained labels, such as the hate target, to enable more comprehensive error analysis \citep[e.g.][]{mathew2021hatexplain,vidgen2021learning}.
Instead of creating a static dataset, \citet{calabrese2021aaa} devise a hate speech-specific data augmentation technique based on simple heuristics to create additional test cases based on model training data.
MHC is the first non-English diagnostic dataset for hate speech detection models.
\vspace{-0.1cm}

\paragraph{Non-English Hate Speech Data}
English is by far the most common language for hate speech datasets, as recent reviews by \citet{vidgen2020directions} and \citet{poletto2021resources} confirm.
Encouragingly, more and more non-English datasets are being created, particularly for shared tasks \citep[e.g.][]{germeval18,ptaszynski2019results,fersini2020ami,zampieri2020semeval,hala2021armi}.
However, very few datasets cover more than one language \citep{ousidhoum2019multilingual, basile2019semeval}, and to our knowledge no dataset covers as many languages as MHC.
\vspace{-0.1cm}

\paragraph{Multilingual Hate Speech Detection}
The scarcity of non-English hate speech datasets has motivated research into few- and zero-shot cross-lingual hate speech detection, i.e. detection with little or no training data in the target language.
However, model performance is generally found to be lacking in such settings \citep{stappen2020crosslingual, leite2020toxic, nozza2021exposing}.
Others have thus explored data augmentation techniques based on machine translation, which yield limited improvements \citep{pamungkas2021joint,wang2021practical}.
Overall, multilingual models trained or fine-tuned directly on the target languages, i.e. in many-shot settings, are still consistently found to perform best \citep{aluru2020multi, pelicon_investigating_2021}.
MHC's functional tests are model-agnostic and can be used to evaluate multilingual hate speech detection models trained on any amount of data.
\section{Conclusion} \label{sec:conclusion}
\vspace{-0.1cm}
In this article, we introduced \textsc{Multilingual HateCheck} (MHC), a suite of functional tests for multilingual hate speech detection models.
MHC expands the English-language \textsc{HateCheck} \citep{rottger2021hatecheck} to ten additional languages:
Arabic, Dutch, French, German, Hindi, Italian, Mandarin, Polish, Portuguese and Spanish.
To our knowledge, MHC covers more languages than any other hate speech dataset.
Across the languages, native-speaking language experts created 36,582 test cases, which provide contrasts between hateful and non-hateful content.
This makes MHC challenging to hate speech detection models and allows for a more effective evaluation of model quality.

We demonstrated MHC's utility as a diagnostic tool by testing a high-performing multilingual transformer model, which was fine-tuned on three widely-used hate speech datasets in three different languages.
MHC revealed the model to be 1) overly sensitive to key words and key phrases, 2) biased in its target coverage and 3) error-prone and inconsistent in cross-lingual transfer, in both zero- and many-shot settings.

So far, hate speech research has primarily focused on English-language content and thus neglected billions of non-English speakers across the world.
We hope that MHC can contribute to closing this language gap and that by diagnosing specific model weaknesses across languages it can support the development of better multilingual hate speech detection models in the future.

\section*{Acknowledgments}
This research was commissioned from Rewire by Google's Jigsaw team.
All authors worked on this project in their capacity as researchers at Rewire.
We thank all annotators and language experts for their work, and all reviewers for their constructive feedback.

\bibliography{custom}

\ifdefined\DeclarePrefChars\DeclarePrefChars{'’-}\else\fi
\begin{thebibliography}{59}
\expandafter\ifx\csname natexlab\endcsname\relax\def\natexlab#1{#1}\fi

\bibitem[{Aluru et~al.(2020)Aluru, Mathew, Saha, and
  Mukherjee}]{aluru2020multi}
Sai~Saketh Aluru, Binny Mathew, Punyajoy Saha, and Animesh Mukherjee. 2020.
\newblock \href {https://doi.org/10.1007/978-3-030-67670-4_26} {A deep dive
  into multilingual hate speech classification}.
\newblock In \emph{Machine Learning and Knowledge Discovery in Databases.
  Applied Data Science and Demo Track: European Conference, ECML PKDD 2020,
  Ghent, Belgium, September 14–18, 2020, Proceedings, Part V}, page
  423–439, Berlin, Heidelberg. Springer-Verlag.

\bibitem[{Banerjee et~al.(2021)Banerjee, Sarkar, Agrawal, Saha, and
  Das}]{banerjee2021exploring}
Somnath Banerjee, Maulindu Sarkar, Nancy Agrawal, Punyajoy Saha, and Mithun
  Das. 2021.
\newblock Exploring transformer based models to identify hate speech and
  offensive content in english and indo-aryan languages.
\newblock \emph{arXiv preprint arXiv:2111.13974}.

\bibitem[{Banko et~al.(2020)Banko, MacKeen, and Ray}]{banko2020unified}
Michele Banko, Brendon MacKeen, and Laurie Ray. 2020.
\newblock \href {https://www.aclweb.org/anthology/2020.alw-1.16} {A {{Unified
  Taxonomy}} of {{Harmful Content}}}.
\newblock In \emph{Proceedings of the {{Fourth Workshop}} on {{Online Abuse}}
  and {{Harms}}}, pages 125--137. {Association for Computational Linguistics}.

\bibitem[{Barbieri et~al.(2021)Barbieri, Anke, and
  Camacho-Collados}]{barbieri2021xlmt}
Francesco Barbieri, Luis~Espinosa Anke, and Jose Camacho-Collados. 2021.
\newblock {XLM-T}: A multilingual language model toolkit for twitter.
\newblock \emph{arXiv preprint arXiv:2104.12250}.

\bibitem[{Basile et~al.(2019)Basile, Bosco, Fersini, Debora, Patti, Pardo,
  Rosso, Sanguinetti et~al.}]{basile2019semeval}
Valerio Basile, Cristina Bosco, Elisabetta Fersini, Nozza Debora, Viviana
  Patti, Francisco Manuel~Rangel Pardo, Paolo Rosso, Manuela Sanguinetti,
  et~al. 2019.
\newblock Semeval-2019 task 5: Multilingual detection of hate speech against
  immigrants and women in twitter.
\newblock In \emph{13th International Workshop on Semantic Evaluation}, pages
  54--63. Association for Computational Linguistics.

\bibitem[{Beizer(1995)}]{beizer1995black}
Boris Beizer. 1995.
\newblock \emph{Black-box testing: techniques for functional testing of
  software and systems}.
\newblock John Wiley \& Sons, Inc.

\bibitem[{Bender and Friedman(2018)}]{bender2018data}
Emily~M. Bender and Batya Friedman. 2018.
\newblock \href {https://doi.org/10.1162/tacl_a_00041} {Data statements for
  natural language processing: Toward mitigating system bias and enabling
  better science}.
\newblock \emph{Transactions of the Association for Computational Linguistics},
  6:587--604.

\bibitem[{Bosco et~al.(2018)Bosco, Felice, Poletto, Sanguinetti, and
  Maurizio}]{bosco2018overview}
Cristina Bosco, Dell'Orletta Felice, Fabio Poletto, Manuela Sanguinetti, and
  Tesconi Maurizio. 2018.
\newblock Overview of the evalita 2018 hate speech detection task.
\newblock In \emph{EVALITA 2018-Sixth Evaluation Campaign of Natural Language
  Processing and Speech Tools for Italian}, volume 2263, pages 1--9. CEUR.

\bibitem[{Calabrese et~al.(2021)Calabrese, Bevilacqua, Ross, Tripodi, and
  Navigli}]{calabrese2021aaa}
Agostina Calabrese, Michele Bevilacqua, Bj{\"o}rn Ross, Rocco Tripodi, and
  Roberto Navigli. 2021.
\newblock Aaa: Fair evaluation for abuse detection systems wanted.
\newblock In \emph{13th ACM Web Science Conference 2021}, pages 243--252.

\bibitem[{Capozzi et~al.(2019)Capozzi, Lai, Basile, Poletto, Sanguinetti,
  Bosco, Patti, Ruffo, Musto, Polignano et~al.}]{capozzi2019computational}
Arthur~TE Capozzi, Mirko Lai, Valerio Basile, Fabio Poletto, Manuela
  Sanguinetti, Cristina Bosco, Viviana Patti, Giancarlo Ruffo, Cataldo Musto,
  Marco Polignano, et~al. 2019.
\newblock Computational linguistics against hate: Hate speech detection and
  visualization on social media in the" contro l’odio" project.
\newblock In \emph{6th Italian Conference on Computational Linguistics, CLiC-it
  2019}, volume 2481, pages 1--6. CEUR-WS.

\bibitem[{Conneau et~al.(2020)Conneau, Khandelwal, Goyal, Chaudhary, Wenzek,
  Guzm{\'a}n, Grave, Ott, Zettlemoyer, and Stoyanov}]{conneau2020xlmr}
Alexis Conneau, Kartikay Khandelwal, Naman Goyal, Vishrav Chaudhary, Guillaume
  Wenzek, Francisco Guzm{\'a}n, Edouard Grave, Myle Ott, Luke Zettlemoyer, and
  Veselin Stoyanov. 2020.
\newblock \href {https://doi.org/10.18653/v1/2020.acl-main.747} {Unsupervised
  cross-lingual representation learning at scale}.
\newblock In \emph{Proceedings of the 58th Annual Meeting of the Association
  for Computational Linguistics}, pages 8440--8451, Online. Association for
  Computational Linguistics.

\bibitem[{Davidson et~al.(2019)Davidson, Bhattacharya, and
  Weber}]{davidson2019racialbias}
Thomas Davidson, Debasmita Bhattacharya, and Ingmar Weber. 2019.
\newblock \href {https://doi.org/10.18653/v1/W19-3504} {Racial bias in hate
  speech and abusive language detection datasets}.
\newblock In \emph{Proceedings of the Third Workshop on Abusive Language
  Online}, pages 25--35, Florence, Italy. Association for Computational
  Linguistics.

\bibitem[{Davidson et~al.(2017)Davidson, Warmsley, Macy, and
  Weber}]{davidson2017automated}
Thomas Davidson, Dana Warmsley, Michael Macy, and Ingmar Weber. 2017.
\newblock Automated hate speech detection and the problem of offensive
  language.
\newblock In \emph{Proceedings of the 11th International AAAI Conference on Web
  and Social Media}, pages 512--515. Association for the Advancement of
  Artificial Intelligence.

\bibitem[{Dixon et~al.(2018)Dixon, Li, Sorensen, Thain, and
  Vasserman}]{dixon2018measuring}
Lucas Dixon, John Li, Jeffrey Sorensen, Nithum Thain, and Lucy Vasserman. 2018.
\newblock \href {https://doi.org/10.1145/3278721.3278729} {Measuring and
  mitigating unintended bias in text classification}.
\newblock In \emph{Proceedings of the 2018 {{AAAI}}/{{ACM Conference}} on
  {{AI}}, {{Ethics}}, and {{Society}}}, pages 67--73. Association for Computing
  Machinery.

\bibitem[{Fersini et~al.(2020)Fersini, Nozza, and Rosso}]{fersini2020ami}
Elisabetta Fersini, Debora Nozza, and Paolo Rosso. 2020.
\newblock \href {http://ceur-ws.org/Vol-2765/paper161.pdf} {{AMI @
  EVALITA2020}: Automatic misogyny identification}.
\newblock In \emph{{Proceedings of the 7th evaluation campaign of Natural
  Language Processing and Speech tools for Italian (EVALITA 2020)}}, Online.
  CEUR.org.

\bibitem[{Fortuna and Nunes(2018)}]{fortuna2018survey}
Paula Fortuna and S{\'e}rgio Nunes. 2018.
\newblock A survey on automatic detection of hate speech in text.
\newblock \emph{ACM Computing Surveys (CSUR)}, 51(4):1--30.

\bibitem[{Fortuna et~al.(2019)Fortuna, Rocha~da Silva, Soler-Company, Wanner,
  and Nunes}]{fortuna2019hierarchically}
Paula Fortuna, Jo{\~a}o Rocha~da Silva, Juan Soler-Company, Leo Wanner, and
  S{\'e}rgio Nunes. 2019.
\newblock \href {https://doi.org/10.18653/v1/W19-3510} {A
  hierarchically-labeled {P}ortuguese hate speech dataset}.
\newblock In \emph{Proceedings of the Third Workshop on Abusive Language
  Online}, pages 94--104, Florence, Italy. Association for Computational
  Linguistics.

\bibitem[{Gardner et~al.(2020)Gardner, Artzi, Basmov, Berant, Bogin, Chen,
  Dasigi, Dua, Elazar, Gottumukkala, Gupta, Hajishirzi, Ilharco, Khashabi, Lin,
  Liu, Liu, Mulcaire, Ning, Singh, Smith, Subramanian, Tsarfaty, Wallace,
  Zhang, and Zhou}]{gardner2020evaluating}
Matt Gardner, Yoav Artzi, Victoria Basmov, Jonathan Berant, Ben Bogin, Sihao
  Chen, Pradeep Dasigi, Dheeru Dua, Yanai Elazar, Ananth Gottumukkala, Nitish
  Gupta, Hannaneh Hajishirzi, Gabriel Ilharco, Daniel Khashabi, Kevin Lin,
  Jiangming Liu, Nelson~F. Liu, Phoebe Mulcaire, Qiang Ning, Sameer Singh,
  Noah~A. Smith, Sanjay Subramanian, Reut Tsarfaty, Eric Wallace, Ally Zhang,
  and Ben Zhou. 2020.
\newblock \href {https://doi.org/10.18653/v1/2020.findings-emnlp.117}
  {Evaluating models{'} local decision boundaries via contrast sets}.
\newblock In \emph{Findings of the Association for Computational Linguistics:
  EMNLP 2020}, pages 1307--1323, Online. Association for Computational
  Linguistics.

\bibitem[{Garg et~al.(2019)Garg, Perot, Limtiaco, Taly, Chi, and
  Beutel}]{garg2019counterfactual}
Sahaj Garg, Vincent Perot, Nicole Limtiaco, Ankur Taly, Ed~H. Chi, and Alex
  Beutel. 2019.
\newblock \href {https://doi.org/10.1145/3306618.3317950} {Counterfactual
  fairness in text classification through robustness}.
\newblock In \emph{Proceedings of the 2019 AAAI/ACM Conference on AI, Ethics,
  and Society}, AIES '19, page 219–226, New York, NY, USA. Association for
  Computing Machinery.

\bibitem[{Geva et~al.(2019)Geva, Goldberg, and Berant}]{geva2019modeling}
Mor Geva, Yoav Goldberg, and Jonathan Berant. 2019.
\newblock \href {https://doi.org/10.18653/v1/D19-1107} {Are we modeling the
  task or the annotator? an investigation of annotator bias in natural language
  understanding datasets}.
\newblock In \emph{Proceedings of the 2019 Conference on Empirical Methods in
  Natural Language Processing and the 9th International Joint Conference on
  Natural Language Processing (EMNLP-IJCNLP)}, pages 1161--1166, Hong Kong,
  China. Association for Computational Linguistics.

\bibitem[{Kirk et~al.(2021)Kirk, Vidgen, R{\"o}ttger, Thrush, and
  Hale}]{kirk2021hatemoji}
Hannah~Rose Kirk, Bertram Vidgen, Paul R{\"o}ttger, Tristan Thrush, and Scott~A
  Hale. 2021.
\newblock Hatemoji: A test suite and adversarially-generated dataset for
  benchmarking and detecting emoji-based hate.
\newblock \emph{arXiv preprint arXiv:2108.05921}.

\bibitem[{Kurrek et~al.(2020)Kurrek, Saleem, and
  Ruths}]{kurrek2020comprehensive}
Jana Kurrek, Haji~Mohammad Saleem, and Derek Ruths. 2020.
\newblock \href {https://doi.org/10.18653/v1/2020.alw-1.17} {Towards a
  comprehensive taxonomy and large-scale annotated corpus for online slur
  usage}.
\newblock In \emph{Proceedings of the Fourth Workshop on Online Abuse and
  Harms}, pages 138--149, Online. Association for Computational Linguistics.

\bibitem[{Leite et~al.(2020)Leite, Silva, Bontcheva, and
  Scarton}]{leite2020toxic}
Jo{\~a}o~Augusto Leite, Diego Silva, Kalina Bontcheva, and Carolina Scarton.
  2020.
\newblock \href {https://aclanthology.org/2020.aacl-main.91} {Toxic language
  detection in social media for {B}razilian {P}ortuguese: New dataset and
  multilingual analysis}.
\newblock In \emph{Proceedings of the 1st Conference of the Asia-Pacific
  Chapter of the Association for Computational Linguistics and the 10th
  International Joint Conference on Natural Language Processing}, pages
  914--924, Suzhou, China. Association for Computational Linguistics.

\bibitem[{Loshchilov and Hutter(2019)}]{loshchilov2019decoupled}
Ilya Loshchilov and Frank Hutter. 2019.
\newblock \href {https://openreview.net/forum?id=Bkg6RiCqY7} {Decoupled weight
  decay regularization}.
\newblock In \emph{Proceedings of the 7th International Conference on Learning
  Representations}.

\bibitem[{Mandl et~al.(2021)Mandl, Modha, Shahi, Madhu, Satapara, Majumder,
  Sch{\"a}fer, Ranasinghe, Zampieri, Nandini et~al.}]{mandl2021hasoc}
Thomas Mandl, Sandip Modha, Gautam~Kishore Shahi, Hiren Madhu, Shrey Satapara,
  Prasenjit Majumder, Johannes Sch{\"a}fer, Tharindu Ranasinghe, Marcos
  Zampieri, Durgesh Nandini, et~al. 2021.
\newblock Overview of the hasoc subtrack at fire 2021: Hate speech and
  offensive content identification in english and indo-aryan languages.
\newblock \emph{arXiv preprint arXiv:2112.09301}.

\bibitem[{Manerba and Tonelli(2021)}]{manerba2021abusechecklist}
Marta~Marchiori Manerba and Sara Tonelli. 2021.
\newblock \href {https://doi.org/10.18653/v1/2021.woah-1.9} {Fine-grained
  fairness analysis of abusive language detection systems with {C}heck{L}ist}.
\newblock In \emph{Proceedings of the 5th Workshop on Online Abuse and Harms
  (WOAH 2021)}, pages 81--91, Online. Association for Computational
  Linguistics.

\bibitem[{Marinescu(2021)}]{marinescu2021facebook}
Delia Marinescu. 2021.
\newblock Facebook's content moderation language barrier.
\newblock \emph{New America}.

\bibitem[{Mathew et~al.(2021)Mathew, Saha, Yimam, Biemann, Goyal, and
  Mukherjee}]{mathew2021hatexplain}
Binny Mathew, Punyajoy Saha, Seid~Muhie Yimam, Chris Biemann, Pawan Goyal, and
  Animesh Mukherjee. 2021.
\newblock Hatexplain: A benchmark dataset for explainable hate speech
  detection.
\newblock In \emph{Proceedings of the AAAI Conference on Artificial
  Intelligence}, volume~35, pages 14867--14875.

\bibitem[{Mulki and Ghanem(2021)}]{hala2021armi}
Hala Mulki and Bilal Ghanem. 2021.
\newblock \href {https://doi.org/10.1145/3503162.3503178} {Working notes of the
  workshop arabic misogyny identification (armi-2021)}.
\newblock In \emph{Forum for Information Retrieval Evaluation}, FIRE 2021, page
  7–8, New York, NY, USA. Association for Computing Machinery.

\bibitem[{Niven and Kao(2019)}]{niven2019probing}
Timothy Niven and Hung-Yu Kao. 2019.
\newblock \href {https://doi.org/10.18653/v1/P19-1459} {Probing neural network
  comprehension of natural language arguments}.
\newblock In \emph{Proceedings of the 57th Annual Meeting of the Association
  for Computational Linguistics}, pages 4658--4664, Florence, Italy.
  Association for Computational Linguistics.

\bibitem[{Nozza(2021)}]{nozza2021exposing}
Debora Nozza. 2021.
\newblock Exposing the limits of zero-shot cross-lingual hate speech detection.
\newblock In \emph{Proceedings of the 59th Annual Meeting of the Association
  for Computational Linguistics}, Online. Association for Computational
  Linguistics.

\bibitem[{Ousidhoum et~al.(2019)Ousidhoum, Lin, Zhang, Song, and
  Yeung}]{ousidhoum2019multilingual}
Nedjma Ousidhoum, Zizheng Lin, Hongming Zhang, Yangqiu Song, and Dit-Yan Yeung.
  2019.
\newblock \href {https://doi.org/10.18653/v1/D19-1474} {Multilingual and
  multi-aspect hate speech analysis}.
\newblock In \emph{Proceedings of the 2019 Conference on Empirical Methods in
  Natural Language Processing and the 9th International Joint Conference on
  Natural Language Processing (EMNLP-IJCNLP)}, pages 4675--4684, Hong Kong,
  China. Association for Computational Linguistics.

\bibitem[{Pamungkas et~al.(2021)Pamungkas, Basile, and
  Patti}]{pamungkas2021joint}
Endang~Wahyu Pamungkas, Valerio Basile, and Viviana Patti. 2021.
\newblock \href {https://doi.org/10.1016/j.ipm.2021.102544} {A joint learning
  approach with knowledge injection for zero-shot cross-lingual hate speech
  detection}.
\newblock \emph{Information Processing \& Management}, 58(4):102544.

\bibitem[{Pelicon et~al.(2021)Pelicon, Shekhar, Škrlj, Purver, and
  Pollak}]{pelicon_investigating_2021}
Andraž Pelicon, Ravi Shekhar, Blaž Škrlj, Matthew Purver, and Senja Pollak.
  2021.
\newblock \href {https://doi.org/10.7717/peerj-cs.559} {Investigating
  cross-lingual training for offensive language detection}.
\newblock \emph{PeerJ Computer Science}, 7:e559.
\newblock Publisher: PeerJ Inc.

\bibitem[{Poletto et~al.(2021)Poletto, Basile, Sanguinetti, Bosco, and
  Patti}]{poletto2021resources}
Fabio Poletto, Valerio Basile, Manuela Sanguinetti, Cristina Bosco, and Viviana
  Patti. 2021.
\newblock \href {https://doi.org/10.1007/s10579-020-09502-8} {Resources and
  benchmark corpora for hate speech detection: a systematic review}.
\newblock \emph{Language Resources and Evaluation}, 55(2):477--523.

\bibitem[{Ptaszynski et~al.(2019)Ptaszynski, Pieciukiewicz, and
  Dyba{\l}a}]{ptaszynski2019results}
Michal Ptaszynski, Agata Pieciukiewicz, and Pawe{\l} Dyba{\l}a. 2019.
\newblock Results of the poleval 2019 shared task 6: First dataset and open
  shared task for automatic cyberbullying detection in polish twitter.
\newblock \emph{Proceedings of the PolEval 2019 Workshop}, page~89.

\bibitem[{Ribeiro et~al.(2020)Ribeiro, Wu, Guestrin, and
  Singh}]{ribeiro2020beyond}
Marco~Tulio Ribeiro, Tongshuang Wu, Carlos Guestrin, and Sameer Singh. 2020.
\newblock \href {https://doi.org/10.18653/v1/2020.acl-main.442} {Beyond
  accuracy: Behavioral testing of {NLP} models with {C}heck{L}ist}.
\newblock In \emph{Proceedings of the 58th Annual Meeting of the Association
  for Computational Linguistics}, pages 4902--4912, Online. Association for
  Computational Linguistics.

\bibitem[{R{\"o}ttger et~al.(2021{\natexlab{a}})R{\"o}ttger, Vidgen, Hovy, and
  Pierrehumbert}]{rottger2021two}
Paul R{\"o}ttger, Bertie Vidgen, Dirk Hovy, and Janet~B Pierrehumbert.
  2021{\natexlab{a}}.
\newblock Two contrasting data annotation paradigms for subjective nlp tasks.
\newblock \emph{arXiv preprint arXiv:2112.07475}.

\bibitem[{R{\"o}ttger et~al.(2021{\natexlab{b}})R{\"o}ttger, Vidgen, Nguyen,
  Talat, Margetts, and Pierrehumbert}]{rottger2021hatecheck}
Paul R{\"o}ttger, Bertie Vidgen, Dong Nguyen, Zeerak Talat, Helen Margetts, and
  Janet Pierrehumbert. 2021{\natexlab{b}}.
\newblock \href {https://doi.org/10.18653/v1/2021.acl-long.4} {{H}ate{C}heck:
  Functional tests for hate speech detection models}.
\newblock In \emph{Proceedings of the 59th Annual Meeting of the Association
  for Computational Linguistics and the 11th International Joint Conference on
  Natural Language Processing (Volume 1: Long Papers)}, pages 41--58, Online.
  Association for Computational Linguistics.

\bibitem[{Sanguinetti et~al.(2020)Sanguinetti, Comandini, Di~Nuovo, Frenda,
  Stranisci, Bosco, Tommaso, Patti, Russo et~al.}]{sanguinetti2020haspeede}
Manuela Sanguinetti, Gloria Comandini, Elisa Di~Nuovo, Simona Frenda,
  Marco~Antonio Stranisci, Cristina Bosco, Caselli Tommaso, Viviana Patti,
  Irene Russo, et~al. 2020.
\newblock Haspeede 2@ evalita2020: Overview of the evalita 2020 hate speech
  detection task.
\newblock In \emph{EVALITA 2020 Seventh Evaluation Campaign of Natural Language
  Processing and Speech Tools for Italian}, pages 1--9. CEUR.

\bibitem[{Sanguinetti et~al.(2018)Sanguinetti, Poletto, Bosco, Patti, and
  Stranisci}]{sanguinetti2018italian}
Manuela Sanguinetti, Fabio Poletto, Cristina Bosco, Viviana Patti, and Marco
  Stranisci. 2018.
\newblock An {I}talian {T}witter corpus of hate speech against immigrants.
\newblock In \emph{Proceedings of the 11th International Conference on Language
  Resources and Evaluation (LREC 2018)}.

\bibitem[{Sap et~al.(2021)Sap, Swayamdipta, Vianna, Zhou, Choi, and
  Smith}]{sap2021annotators}
Maarten Sap, Swabha Swayamdipta, Laura Vianna, Xuhui Zhou, Yejin Choi, and
  Noah~A. Smith. 2021.
\newblock \href {http://arxiv.org/abs/2111.07997} {Annotators with attitudes:
  How annotator beliefs and identities bias toxic language detection}.

\bibitem[{Schmidt and Wiegand(2017)}]{schmidt2017survey}
Anna Schmidt and Michael Wiegand. 2017.
\newblock \href {https://doi.org/10.18653/v1/W17-1101} {A survey on hate speech
  detection using natural language processing}.
\newblock In \emph{Proceedings of the Fifth International Workshop on Natural
  Language Processing for Social Media}, pages 1--10, Valencia, Spain.
  Association for Computational Linguistics.

\bibitem[{Shah et~al.(2020)Shah, Schwartz, and Hovy}]{shah2020predictive}
Deven~Santosh Shah, H.~Andrew Schwartz, and Dirk Hovy. 2020.
\newblock \href {https://doi.org/10.18653/v1/2020.acl-main.468} {Predictive
  biases in natural language processing models: A conceptual framework and
  overview}.
\newblock In \emph{Proceedings of the 58th Annual Meeting of the Association
  for Computational Linguistics}, pages 5248--5264, Online. Association for
  Computational Linguistics.

\bibitem[{Simonite(2021)}]{simonite2021facebook}
Tom Simonite. 2021.
\newblock Facebook is everywhere; its moderation is nowhere close.
\newblock \emph{Wired}.

\bibitem[{Stappen et~al.(2020)Stappen, Brunn, and
  Schuller}]{stappen2020crosslingual}
Lukas Stappen, Fabian Brunn, and Bj{\"{o}}rn~W. Schuller. 2020.
\newblock \href {http://arxiv.org/abs/2004.13850} {Cross-lingual zero- and
  few-shot hate speech detection utilising frozen transformer language models
  and {AXEL}}.
\newblock \emph{CoRR}, abs/2004.13850.

\bibitem[{Talat(2016)}]{waseem2016you}
Zeerak Talat. 2016.
\newblock \href {https://doi.org/10.18653/v1/W16-5618} {Are you a racist or am
  {I} seeing things? {A}nnotator influence on hate speech detection on
  {T}witter}.
\newblock In \emph{Proceedings of the First Workshop on {NLP} and Computational
  Social Science}, pages 138--142, Austin, Texas. Association for Computational
  Linguistics.

\bibitem[{van Aken et~al.(2018)van Aken, Risch, Krestel, and
  L{\"o}ser}]{van2018challenges}
Betty van Aken, Julian Risch, Ralf Krestel, and Alexander L{\"o}ser. 2018.
\newblock \href {https://doi.org/10.18653/v1/W18-5105} {Challenges for toxic
  comment classification: An in-depth error analysis}.
\newblock In \emph{Proceedings of the 2nd Workshop on Abusive Language Online
  ({ALW}2)}, pages 33--42, Brussels, Belgium. Association for Computational
  Linguistics.

\bibitem[{Vidgen and Derczynski(2020)}]{vidgen2020directions}
Bertie Vidgen and Leon Derczynski. 2020.
\newblock \href {https://doi.org/10.1371/journal.pone.0243300} {Directions in
  abusive language training data, a systematic review: {{Garbage}} in, garbage
  out}.
\newblock \emph{PLOS ONE}, 15(12):e0243300.

\bibitem[{Vidgen et~al.(2020)Vidgen, Hale, Guest, Margetts, Broniatowski,
  Talat, Botelho, Hall, and Tromble}]{vidgen2020detecting}
Bertie Vidgen, Scott Hale, Ella Guest, Helen Margetts, David Broniatowski,
  Zeerak Talat, Austin Botelho, Matthew Hall, and Rebekah Tromble. 2020.
\newblock \href {https://doi.org/10.18653/v1/2020.alw-1.19} {Detecting {E}ast
  {A}sian prejudice on social media}.
\newblock In \emph{Proceedings of the Fourth Workshop on Online Abuse and
  Harms}, pages 162--172, Online. Association for Computational Linguistics.

\bibitem[{Vidgen et~al.(2019)Vidgen, Harris, Nguyen, Tromble, Hale, and
  Margetts}]{vidgen2019challenges}
Bertie Vidgen, Alex Harris, Dong Nguyen, Rebekah Tromble, Scott Hale, and Helen
  Margetts. 2019.
\newblock \href {https://doi.org/10.18653/v1/W19-3509} {Challenges and
  frontiers in abusive content detection}.
\newblock In \emph{Proceedings of the Third Workshop on Abusive Language
  Online}, pages 80--93, Florence, Italy. Association for Computational
  Linguistics.

\bibitem[{Vidgen et~al.(2021)Vidgen, Thrush, Talat, and
  Kiela}]{vidgen2021learning}
Bertie Vidgen, Tristan Thrush, Zeerak Talat, and Douwe Kiela. 2021.
\newblock \href {https://doi.org/10.18653/v1/2021.acl-long.132} {Learning from
  the worst: Dynamically generated datasets to improve online hate detection}.
\newblock In \emph{Proceedings of the 59th Annual Meeting of the Association
  for Computational Linguistics and the 11th International Joint Conference on
  Natural Language Processing (Volume 1: Long Papers)}, pages 1667--1682,
  Online. Association for Computational Linguistics.

\bibitem[{Wang and Banko(2021)}]{wang2021practical}
Cindy Wang and Michele Banko. 2021.
\newblock \href {https://doi.org/10.18653/v1/2021.naacl-industry.16} {Practical
  {Transformer}-based {Multilingual} {Text} {Classification}}.
\newblock In \emph{Proceedings of the 2021 {Conference} of the {North}
  {American} {Chapter} of the {Association} for {Computational} {Linguistics}:
  {Human} {Language} {Technologies}: {Industry} {Papers}}, pages 121--129,
  Online. Association for Computational Linguistics.

\bibitem[{Wiegand et~al.(2019)Wiegand, Ruppenhofer, and
  Kleinbauer}]{wiegand2019detection}
Michael Wiegand, Josef Ruppenhofer, and Thomas Kleinbauer. 2019.
\newblock \href {https://doi.org/10.18653/v1/N19-1060} {{D}etection of abusive
  language: The problem of biased datasets}.
\newblock In \emph{Proceedings of the 2019 Conference of the North {A}merican
  Chapter of the Association for Computational Linguistics: Human Language
  Technologies, Volume 1 (Long and Short Papers)}, pages 602--608, Minneapolis,
  Minnesota. Association for Computational Linguistics.

\bibitem[{Wiegand et~al.(2018)Wiegand, Siegel, and Ruppenhofer}]{germeval18}
Michael Wiegand, Melanie Siegel, and Josef Ruppenhofer. 2018.
\newblock {Overview of the GermEval 2018 Shared Task on the Identification of
  Offensive Language}.
\newblock In \emph{Proceedings of GermEval 2018, 14th Conference on Natural
  Language Processing (KONVENS 2018)}.

\bibitem[{Wolf et~al.(2020)Wolf, Debut, Sanh, Chaumond, Delangue, Moi, Cistac,
  Rault, Louf, Funtowicz, Davison, Shleifer, von Platen, Ma, Jernite, Plu, Xu,
  Le~Scao, Gugger, Drame, Lhoest, and Rush}]{wolf2020transformers}
Thomas Wolf, Lysandre Debut, Victor Sanh, Julien Chaumond, Clement Delangue,
  Anthony Moi, Pierric Cistac, Tim Rault, Remi Louf, Morgan Funtowicz, Joe
  Davison, Sam Shleifer, Patrick von Platen, Clara Ma, Yacine Jernite, Julien
  Plu, Canwen Xu, Teven Le~Scao, Sylvain Gugger, Mariama Drame, Quentin Lhoest,
  and Alexander Rush. 2020.
\newblock \href {https://doi.org/10.18653/v1/2020.emnlp-demos.6} {Transformers:
  State-of-the-art natural language processing}.
\newblock In \emph{Proceedings of the 2020 Conference on Empirical Methods in
  Natural Language Processing: System Demonstrations}, pages 38--45, Online.
  Association for Computational Linguistics.

\bibitem[{Wu et~al.(2019)Wu, Ribeiro, Heer, and Weld}]{wu2019errudite}
Tongshuang Wu, Marco~Tulio Ribeiro, Jeffrey Heer, and Daniel Weld. 2019.
\newblock \href {https://doi.org/10.18653/v1/P19-1073} {{E}rrudite: Scalable,
  reproducible, and testable error analysis}.
\newblock In \emph{Proceedings of the 57th Annual Meeting of the Association
  for Computational Linguistics}, pages 747--763, Florence, Italy. Association
  for Computational Linguistics.

\bibitem[{Zampieri et~al.(2019)Zampieri, Malmasi, Nakov, Rosenthal, Farra, and
  Kumar}]{zampieri2019predicting}
Marcos Zampieri, Shervin Malmasi, Preslav Nakov, Sara Rosenthal, Noura Farra,
  and Ritesh Kumar. 2019.
\newblock \href {https://doi.org/10.18653/v1/N19-1144} {Predicting the type and
  target of offensive posts in social media}.
\newblock In \emph{Proceedings of the 2019 Conference of the North {A}merican
  Chapter of the Association for Computational Linguistics: Human Language
  Technologies, Volume 1 (Long and Short Papers)}, pages 1415--1420,
  Minneapolis, Minnesota. Association for Computational Linguistics.

\bibitem[{Zampieri et~al.(2020)Zampieri, Nakov, Rosenthal, Atanasova,
  Karadzhov, Mubarak, Derczynski, Pitenis, and
  \c{C}\"{o}ltekin}]{zampieri2020semeval}
Marcos Zampieri, Preslav Nakov, Sara Rosenthal, Pepa Atanasova, Georgi
  Karadzhov, Hamdy Mubarak, Leon Derczynski, Zeses Pitenis, and
  \c{C}a\u{g}r{\i} \c{C}\"{o}ltekin. 2020.
\newblock {SemEval-2020 Task 12: Multilingual Offensive Language Identification
  in Social Media (OffensEval 2020)}.
\newblock In \emph{Proceedings of SemEval}.

\end{thebibliography}

\clearpage
\appendix

\section{Example MHC Test Cases} \label{app: examples}

See Table \ref{fig: examples} on the next page.

\begin{table*}
\small
\centering
\resizebox{\textwidth}{!}{%
\renewcommand{\arraystretch}{1.33}
\begin{tabular}{p{0.08\textwidth}m{0.35\textwidth}C{0.1\textwidth}|C{0.1\textwidth}m{0.4\textwidth}}
\toprule & \multirow{1}{*}{\textbf{Functionality}} & \multirow{1}{*}{\textbf{Gold Label}} & \multirow{1}{*}{\textbf{Language}} & \multirow{1}{*}{\textbf{Example}}  \\
\midrule

\multirow{4}{*}{\rotatebox[origin=c]{0}{\parbox[c]{1.3cm}{Derogation}}}
& \textbf{F1}: Expression of strong negative emotions (explicit) & \hlred{ hateful} & FR & Il n'y a rien de plus horrible qu'une femme. \\
 & \textbf{F2}: Description using very negative attributes (explicit) & \hlred{ hateful } & HI & duniya mai sabse zyada murkh bas gays hi hote hain. \\
& \textbf{F3}: Dehumanisation (explicit) & \hlred{ hateful } & NL &   Gehandicapten zijn een plaag. \\
& \textbf{F4}: Implicit derogation & \hlred{ hateful } & PL & Życie każdego owada jest cenniejsze niż życie kobiety. \\
\midrule

\multirow{2}{*}{\rotatebox[origin=c]{0}{\parbox[c]{1.2cm}{Threat. language}}}
& \textbf{F5}: Direct threat & \hlred{ hateful }
& ZH & \begin{CJK*}{UTF8}{gbsn}只能给女人死刑才能解决问题。\end{CJK*}\\
& \textbf{F6}: Threat as normative statement & \hlred{ hateful } & ES & En un mundo justo, los negros como ustedes estarían todos muertos.\\
\midrule

\multirow{1}{*}{\rotatebox[origin=c]{0}{\parbox[c]{1.3cm}{Slurs}}}
& \textbf{F7}: Hate expressed using slur & \hlred{ hateful } & PT & Anormais simplesmente não conseguem ajudar a si mesmos.\\
\midrule

\multirow{2}{10pt}{\rotatebox[origin=c]{0}{\parbox[c]{1.2cm}{Profanity usage}}} 
& \textbf{F8}: Hate expressed using profanity & \hlred{ hateful } & IT & Tipico di un trans essere un pezzo di merda.\\
& \textbf{F9}: Non-hateful use of profanity & \hlgreen{ non-hate } & AR & \<لازم احاول قد مقدر ما بقاش حمار.>\\
\midrule

\multirow{2}{*}{\rotatebox[origin=c]{0}{\parbox[c]{1.3cm}{Pronoun reference}}}
& \textbf{F10}: Hate expressed through reference in subsequent clauses & \hlred{ hateful } & HI & wo shudra insaan dikhte hai par hai wo parijeevi hi.\\
 & \textbf{F11}: Hate expressed through reference in subsequent sentences & \hlred{ hateful }
 & PT & É obvio que eles são estúpidios, afiinal são gays\\
\midrule

\multirow{2}{*}{\rotatebox[origin=c]{0}{\parbox[c]{1.3cm}{Negation}}}
& \textbf{F12}: Hate expressed using negated positive statement & \hlred{ hateful } & DE & Behinderte sind uns grundsätzlich nicht ebenbürtig.\\
& \textbf{F13}: Non-hate expressed using negated hateful statement & \hlgreen{ non-hate } & PL & Nie stójmy bezczynnie, gdy niepełnosprawni są mordowani.\\
\midrule

\multirow{2}{*}{\rotatebox[origin=c]{0}{\parbox[c]{1.3cm}{Phrasing}}}
& \textbf{F14}: Hate phrased as a question & \hlred{ hateful }
& IT & Sbaglio a pensare che questi disabili debbano essere condannati a morte?\\
& \textbf{F15}: Hate phrased as an opinion & \hlred{ hateful } & ES & Si me preguntas, los negros como tú no son mejores que las ratas.\\
\midrule

\multirow{2}{*}{\rotatebox[origin=c]{0}{\parbox[c]{1.3cm}{Non-hateful group identifier}}} 
& \textbf{F16}: Neutral statements using protected group identifiers & \hlgreen{ non-hate }
& ZH & \begin{CJK*}{UTF8}{gbsn}我和两个跨性别者住在一起。\end{CJK*}\\
& \textbf{F17}: Positive statements using protected group identifiers & \hlgreen{ non-hate }
& NL & Homoseksuelen horen thuis in de regering. \\
\midrule

\multirow{2}{*}{\rotatebox[origin=c]{0}{\parbox[c]{1.3cm}{Counter speech}}}
& \textbf{F18}: Denouncements of hate that quote it & \hlgreen{ non-hate } & AR & \<ايه اللي خلاك تقول انك بتتقرف من النساء.> \\
& \textbf{F19}: Denouncements of hate that make direct reference to it & \hlgreen{ non-hate } & HI & logo ko kinnar bulana bilkul theek nahi hai.\\
\midrule

\multirow{3}{*}{\rotatebox[origin=c]{0}{\parbox[c]{1.3cm}{Abuse against non-protected targets}}}
& \textbf{F20}: Abuse targeted at objects & \hlgreen{ non-hate } & IT & Questa serie di esercizi è così nauseante.\\
& \textbf{F21}: Abuse targeted at individuals (not as member of a protected group) & \hlgreen{ non-hate } & PL & Nienawidzę cię.\\
& \textbf{F22}: Abuse targeted at non-protected groups (e.g. professions) & \hlgreen{ non-hate }
& DE & Ich kann Politiker wie dich wirklich nicht ausstehen.\\
\midrule

\multirow{12}{*}{\rotatebox[origin=c]{0}{\parbox[c]{1.3cm}{Spelling variations}}}
& \textbf{F23}: Swaps of adjacent characters & \hlred{ hateful } & NL & Zulke transgenders moeten worden geëxecueterd.\\
& \textbf{F24}: Missing characters & \hlred{ hateful } & FR & Typique d'un homo d'être à ce point stupde.\\
& \textbf{F25}: Missing word boundaries & \hlred{ hateful } & PT & Essas mulheresme dão tanto nojo.\\
& \textbf{F26}: Added spaces between chars & \hlred{ hateful } & ES & Quiero m a t a r a todos los indígenas.\\
& \textbf{F27}: Leet speak spellings & \hlred{ hateful }
& HI & keval aurat hi 1tni badi murkh ho sakta hai. \\

& \textbf{F28}: AR: Latin char. replacement & \hlred{ hateful } & AR & \<الص يا يهود.>5\<انا بكرهكم >\\
& \textbf{F29}: AR: Repeated characters & \hlred{ hateful } & AR & \<ما حد فيكو ينزل ضررررررب بالمثلين دول بالمطوة.>\\
& \textbf{F30}: AR: Arabizi (Arabic chat alphabet) & \hlred{ hateful } & AR & Alanwa3 dee mal el yahood lazm ytnafth feha el23dam.\\
& \textbf{F31}: AR: Accepted alt. spellings & \hlred{ hateful } & AR & \<النساء لازمهم حبس في الاآفاص.> \\
& \textbf{F32}: ZH: Homophone char. replacement & \hlred{ hateful } & ZH & \begin{CJK*}{UTF8}{gbsn}我想沙死所有黑人。\end{CJK*}\\
& \textbf{F33}: ZH: Character decomposition & \hlred{ hateful } & ZH & \begin{CJK*}{UTF8}{gbsn}这些外国人就该闭上他们的犭句嘴。\end{CJK*}\\
& \textbf{F34}: ZH: Pinyin spelling & \hlred{ hateful } & ZH & \begin{CJK*}{UTF8}{gbsn}所有女人都去si。\end{CJK*}\\
\bottomrule
\end{tabular}}
\caption{\label{fig: examples}
Example test cases for each of the 34 functional tests in MHC.
Examples were selected at random.
}
\end{table*}

\section{Protected Groups in MHC} \label{app: prot groups}

MHC covers seven protected groups per language.

\textbf{Arabic}:
women /{\small\<نساء>},
trans people / {\small\<متحولين جنسيا>},
gay people / {\small\<مثلين>},
black people / {\small\<سود>},
Jews / {\small\<يهود>},
disabled people / {\small\<معاقيين>},
immigrants / {\small\<لاجئين>}

\textbf{Dutch}:
women / vrouwen,
trans people / transgenders,
gay people / homoseksuelen,
black people / zwarte personen,
disabled people / gehandicapten,
immigrants / immigranten,
Muslims / Moslims

\textbf{French}:
women / femmes,
trans people / trans,
gay people / homos,
black people / blacks,
disabled people / handicapés,
Muslims / musulmans,
refugees / réfugiés

\textbf{German}:
women / Frauen,
trans people / Transmenschen,
gay people / Schwule,
black people / Schwarze,
disabled people / Behinderte
Jews / Juden,
refugees / Flüchtlinge

\textbf{Hindi}:
women / aurato,
trans people / hijro,
gay people / gays,
North-East Indians / assamese,
Muslims / musalmano,
Lower Caste / shudro,
Black people / negro

\textbf{Italian}:
women / donne,
trans people / trans,
gay people / gay,
black people / persone di colore,
disabled people / disabili,
Muslims / musulmani,
immigrants / immigrati

\textbf{Mandarin}:
women / \begin{CJK*}{UTF8}{gbsn}女人\end{CJK*},
trans people / \begin{CJK*}{UTF8}{gbsn}跨性别者\end{CJK*},
gay people / \begin{CJK*}{UTF8}{gbsn}同性恋\end{CJK*},
black people / \begin{CJK*}{UTF8}{gbsn}黑人\end{CJK*},
disabled people / \begin{CJK*}{UTF8}{gbsn}残疾人\end{CJK*},
Muslims / \begin{CJK*}{UTF8}{gbsn}穆斯林\end{CJK*},
foreigners / \begin{CJK*}{UTF8}{gbsn}外国人\end{CJK*}

\textbf{Polish}:
women / kobiety,
trans people / osoby transpłciowe,
gay people / geje,
Asian people / azjaci,
disabled people / niepełnosprawni,
Jews / Żydzi,
immigrants / imigranci

\textbf{Portuguese}:
women / mulheres,
black people / negros,
gay people / gays,
trans people / pessoas trans,
indigenous people / indígenas,
Jews / judeus,
disabled people / deficientes

\textbf{Spanish}:
women / mujeres,
black people / negros,
gay people / gays,
trans people / trans,
indigenous people / indígenas,
Jews / judíos,
disabled people / discapacitados

\section{Data Statement} \label{app: data statement}

Following \newcite{bender2018data}, we provide a data statement, which documents the generation and provenance of test cases in \textsc{MHC}.

\paragraph{A. CURATION RATIONALE}
The goal of our research was to construct MHC, a multilingual suite of functional tests for hate speech detection models.
For this purpose, our team of native-speaking language experts generated a total of 36,582 short text documents in ten different languages, by hand and by using simple templates for group identifiers and slurs (\S\ref{subsec: generating test cases}).
Each document corresponds to one functional test and a binary gold standard label (hateful or non-hateful).

\paragraph{B. LANGUAGE VARIETY}
MHC covers ten languages: Arabic, Dutch, French, German, Hindi, Italian, Mandarin, Polish, Portuguese and Spanish.

\paragraph{C. SPEAKER DEMOGRAPHICS}
All test cases across the ten languages in MHC were hand-crafted by native-speaking language experts -- one per language.
All ten had previously worked on hate speech as researchers and/or annotators.
Six out of ten experts identify as women, the rest as men. Four out of ten identify as non-White.

\paragraph{D. ANNOTATOR DEMOGRAPHICS}
More than 120 annotators provided annotations on MHC, with at least 12 annotators per language.
Annotators were recruited on Appen, a crowdworking provider.
Appen gave no demographic information beyond guaranteeing that annotators were native speakers of the languages in which they completed their work.
In setting up the annotation task and communicating with annotators, we followed guidance for protecting and monitoring annotator well-being provided by \citet{vidgen2019challenges}.

\paragraph{E. SPEECH SITUATION}
All test cases were created between November 2021 and January 2022.

\paragraph{F. TEXT CHARACTERISTICS}
The composition of the dataset is described in detail in \S\ref{subsec: func in test suite} and \S\ref{subsec: generating test cases} of the article.

\section{Annotator Disagreement on MHC} \label{app: disagreement}
Annotator disagreement on MHC (Table \ref{fig: disagreement}) is higher than on the original \textsc{HateCheck} \citep{rottger2021hatecheck}, where four out of five annotators agreed on the gold label in 99.4\% of cases.
There is a lot of variation in disagreement across languages, with most having less than 5\% disagreement, and only Mandarin and French more than 10\%.
Upon review, our language experts found that the vast majority of disagreements stemmed from annotator error, where annotators failed to apply the explicit, prescriptive annotation guidelines they received.
For example, hate and more general abuse were often confused, and abuse against non-protected targets was often labelled as hateful.
Therefore, we did not exclude any cases from MHC.
To enable further analysis and data filtering, we provide annotator labels with the test suite and mark up cases and templates where there is disagreement between the annotator majority labels and the gold labels from our language experts.

\begin{table}[H]
\centering
\small
\begin{tabular}{l|ll}
\toprule
  \textbf{Language} & \textbf{\% Disagreement} & \textbf{n Disagreement} \\ 
\midrule
    \textbf{Arabic / AR} &   7.05 & 252 \\
    \textbf{Dutch / NL} &   9.61 & 362    \\
    \textbf{French / FR} &   21.22 &       789  \\
    \textbf{German / DE} &   4.20 &       153  \\
    \textbf{Hindi / HI} &   4.88 &       174   \\
    \textbf{Italian / IT} &   0.73 &       27  \\
    \textbf{Mandarin / ZH} &   11.48 &       388  \\
    \textbf{Polish / PL} &   2.04 &       78  \\
    \textbf{Portuguese / PT} &   4.12 &       152 \\
    \textbf{Spanish / ES} &   2.40 &       90 \\
\bottomrule
\end{tabular}
\caption{\label{fig: disagreement}
Proportion of entries and absolute number of entries where at least 2/3 annotators disagreed with the expert gold label, for each language in MHC.}
\end{table}

\section{Datasets for Model Fine-Tuning} \label{app: training data}

\subsection{\citet{sanguinetti2020haspeede} Italian Data}

\paragraph{Sampling}
The authors compiled 8,100 tweets sampled using keywords.
4,000 tweets come from HaSpeeDe 2018 \citep{bosco2018overview}, which in turn originates in the \citet{sanguinetti2018italian} dataset.
The other 4,100 tweets were collected as part of the Italian hate speech monitoring project "Contro l’Odio" \citep{capozzi2019computational}.

\paragraph{Annotation}
The \citet{sanguinetti2018italian} tweets were annotated in two phases, first by expert annotators, then by crowdworkers from CrowdFlower.
Each tweet was annotated by two to three annotators for six attributes: \textit{hate speech}, \textit{aggressiveness}, \textit{offensiveness}, \textit{irony}, \textit{stereotype}, and \textit{intensity}.
For inter-annotator agreement, the authors report a Krippendorff's Alpha of 38\% for CrowdFlower, and a Cohen's Kappa of 45\% for the expert annotators.
The "Contro l'Odio" tweets were annotated by crowdworkers, but inter-annotator agreement was not reported. \citep{sanguinetti2020haspeede}.

\paragraph{Data}
We use all 8,100 tweets (41.8\% hate).

\paragraph{Definition of Hate Speech}
"Language that spreads, incites, promotes or justifies hatred or violence towards the given target, or a message that aims at dehumanizing, delegitimizing, hurting or intimidating the target. The targets are Immigrants, Muslims, and Roma groups, or individual members of such groups."

\subsection{\citet{fortuna2019hierarchically} Portuguese Data}

\paragraph{Sampling}
\citet{fortuna2019hierarchically} initially collected 42,930 tweets based on a search of 29 user profiles, 19 keywords and ten hashtags.
They then filtered the tweets, keeping only Portuguese-language tweets, and removing duplicates and retweets, resulting in 33,890 tweets.
Finally, they set a cap of a maximum of 200 tweets per search method, to create the final dataset of 5,668 tweets.

\paragraph{Annotation}
All tweets in the dataset were annotated as either hateful or non-hateful by 18 non-expert Portuguese native speakers were hired.
Each tweet was annotated by three annotators, and inter-annotator agreement was low, with a Cohen's Kappa of 0.17.

\paragraph{Data}
We use all 5,668 tweets (31.5\% hate).

\paragraph{Definition of Hate Speech}
"Hate speech is language that attacks or diminishes, that incites violence or hate against groups, based on specific characteristics such as physical appearance, religion, descent, national or ethnic origin, sexual orientation, gender identity or other, and it can occur with different linguistic styles, even in subtle forms or when humour is used."

\subsection{\citet{basile2019semeval} Spanish Data}

\paragraph{Sampling}
Tweets were sampled using three methods:
1) monitoring potential victims of hate accounts,
2) retrieving tweets from the history of identified haters, and
3) retrieving tweets using neutral and derogatory keywords, polarising hashtags, and stems.
This yielded 19,600 tweets, of which 6,600 are in Spanish and the rest in English.

\paragraph{Annotation}
The dataset was annotated for three attributes: \textit{hate speech}, \textit{target range} (individuals or groups), and \textit{aggressiveness}.
First, all data was annotated by at least three Figure Eight crowdworkers.
Inter-annotator agreement on Spanish \textit{hate speech} was high, with a Cohen's Kappa of 0.89.
Second, two experts annotated each tweet.
The final label was assigned based on majority vote across the crowd and expert annotators.

\paragraph{Data}
We use all 6,600 Spanish tweets, of which 41.5\% are labelled as hateful.

\paragraph{Definition of Hate Speech}
"Any communication that disparages a person or a group on the basis of some characteristic such as race, color, ethnicity, gender, sexual orientation, nationality, religion, or other characteristics."

\subsection{Pre-Processing}
Before using the datasets for fine-tuning, we remove newline and tab characters.
We replace URLs and user mentions with [URL] and [USER] tokens.

\section{XLM-T Model Comparison} \label{app: model comparison}

We denote the three XLM-T models trained on Italian \citet{sanguinetti2020haspeede}, Portuguese \citet{fortuna2019hierarchically} and Spanish \citet{basile2019semeval} as XLM-IT, XLM-PT and XLM-ES respectively.
XTC denotes the XLM-T model trained on the combination of all three datasets, for which we report results in the main body of this article.
XTC generally outperforms the monolingual models when compared on the respective held-out test sets (Table~\ref{tab: other_models_heldout}) as well as MHC (Table \ref{tab: other_models_mhc}).

\begin{table}[h]
    \small
    \centering
    \begin{tabular}{l|cccc}
\toprule
   \textbf{Dataset} &  \textbf{XLM-IT} &  \textbf{XLM-PT} & \textbf{XLM-ES} & \textbf{XTC}  \\
\midrule
    IT &   73.2 & - & - & \textbf{76.3} \\
    PT & - & \textbf{75.3} & - & 73.3 \\
    ES & - & - & 84.0 & \textbf{84.7} \\
\bottomrule
\end{tabular}
\caption{Macro F1 for each fine-tuned model on its respective test set and for XTC on all test sets.}
\label{tab: other_models_heldout}
\end{table}

\begin{table}[h]
    \small
    \centering
    
    \begin{tabular}{l|cccc}
\toprule
   \textbf{Lang.} &  \textbf{XLM-IT} &  \textbf{XLM-PT} & \textbf{XLM-ES} & \textbf{XTC}  \\
\midrule
    AR &   51.3 &      45.8 &   51.4 & \textbf{59.4} \\
    NL &   59.9 &      49.5 &   59.6 & \textbf{66.7} \\
    FR &   57.5 &      50.5 &   62.2 & \textbf{67.6} \\
    DE &   62.1 &      46.9 &   59.5 & \textbf{68.9} \\
    HI &   48.2 &      44.4 &   47.4 & \textbf{58.1} \\
    IT &   53.6 &      47.0 &   54.6 & \textbf{69.6} \\
    ZH &   61.8 &      42.7 &   53.2 & \textbf{71.5} \\
    PL &   57.5 &      49.2 &   58.2 & \textbf{66.2} \\
    PT &   58.6 &      64.2 &   56.0 & \textbf{68.5} \\
    ES &   60.0 &      50.1 &   64.4 & \textbf{69.5} \\

\bottomrule
\end{tabular}
\caption{\label{tab: other_models_mhc}
Macro F1 across languages on MHC for each of our fine-tuned models.}
\end{table}

\section{XLM-T Model Details} \label{app: hyperparameters}

\paragraph{Model Architecture}
We implemented XLM-T model \citep{barbieri2021xlmt} using the \texttt{transformers} Python library \citep{wolf2020transformers}.
XLM-T is an XLM-R \citep{conneau2020xlmr} model pre-trained on an additional 198 million Twitter posts in over 30 languages.
It has 12 layers, a hidden layer size of 768, 12 attention heads and a total of 278 million parameters.
For sequence classification, we added a linear layer with softmax output.

\paragraph{Fine-Tuning}
All models use unweighted cross-entropy loss and the AdamW optimiser \citep{loshchilov2019decoupled} with a 5e-5 learning rate and a 0.01 weight decay.
For regularisation, we set a 10\% dropout probability, and for batch size we use 32. For each model, we train for 50 epochs, with an early stopping strategy with a patience of 5 epochs, with respect to improvements in the binary F1-score on the validation split. We store the checkpoint with the highest binary F1-score and use it as our final model.

\paragraph{Computation}
We ran all computations on an AWS  "g4dn.2xlarge" server equipped with one NVIDIA T4 GPU card.
The average wall time for each each training step was around 3 seconds.

\paragraph{Model Access}
We make the XTC model available for download on \href{https://huggingface.co/Rewire/XTC}{HuggingFace}.

\section{Google Perspective Results} \label{app: perspective results}

We test Perspective's "identity attack" attribute and convert the percentage score to a binary label using a 50\% cutoff.
Testing was done in February 2022.

On the held-out test sets for Italian \citep{sanguinetti2020haspeede} and Portuguese \citep{fortuna2019hierarchically}, Perspective scored 70.7 and 64.1 macro F1.
Perspective is outperformed on both languages by XTC, which scored 76.3 and 84.7 (Table \ref{tab: other_models_heldout}).

On MHC, for the three languages it supports, Perspective (Table \ref{table: perspective on mhc}) performs worse than XTC (Table \ref{fig: f1_by_label}) in terms of macro F1 for Italian and Portuguese, and around equally well for German.

\begin{table}[H]
\centering
\small
\begin{tabular}{l|cc|c}
\toprule
  \textbf{Language} & \textbf{F1-h} & \textbf{F1-nh} & \textbf{Macro F1} \\
\midrule
    \textbf{German / DE} &   84.1  &     54.9 &    69.5 \\
    \textbf{Italian / IT} &   69.6 &      61.2 &    65.4 \\
    \textbf{Portuguese / PT} &   84.2 &       47.6 &   65.9 \\
\bottomrule
\end{tabular}
\caption{
Performance of the Perspective API across the three languages it supports in MHC. F1 score for \textbf{h}ateful and \textbf{n}on-\textbf{h}ateful cases, and overall macro F1 score.}
\label{table: perspective on mhc}
\end{table}

\end{document}